\documentclass[10pt,twocolumn,letterpaper]{article}

\usepackage{cvpr}              
\usepackage{pifont}
\usepackage{graphicx}
\usepackage{colortbl}
\usepackage{arydshln}
\usepackage{multirow}
\usepackage{amssymb}
\usepackage{bm}
\usepackage{calc}
\usepackage{subcaption}
\usepackage{Yin}
\usepackage{hhline}
\usepackage{tikz}
\usepackage{tcolorbox}
\usepackage{colortbl} 
\usepackage{algorithmic}
\usepackage{algorithm}

\makeatletter
\renewcommand{\p@subtable}{\thetable}

\makeatother

\definecolor{cvprblue}{rgb}{0.21,0.49,0.74}
\definecolor{unitcolor3}{RGB}{240,250,254}
\usepackage[pagebackref,breaklinks,colorlinks,allcolors=cvprblue]{hyperref}
\definecolor{urlcolor}{RGB}{236,0,140}

\def\paperID{XXXX} 
\def\confName{CVPR}
\def\confYear{2026}

\hypersetup{
    colorlinks=true,
    urlcolor=urlcolor,   
    linkcolor=red,
}

\title{PCA-Seg: Revisiting  Cost Aggregation for Open-Vocabulary Semantic and Part Segmentation}

\author{Jianjian~Yin$^{1,5}$~~~Tao~Chen$^{1,5}$\thanks{Corresponding author.} ~~~Yi~Chen$^2$~~~Gensheng~Pei$^3$~~~Xiangbo~Shu$^1$~~~Yazhou~Yao$^{1,5*}$~~~Fumin Shen$^4$\vspace{.5mm}\\
$^1$Nanjing University of Science and Technology~~~~~~~~~$^2$Nanjing Normal University\\
$^3$Department of Electrical and Computer Engineering,~~ Sungkyunkwan University\\
$^4$University of Electronic Science and Technology of China \\
$^5$State Key Laboratory of Intelligent Manufacturing of Advanced Construction Machinery \\
\small{\url{https://github.com/NUST-Machine-Intelligence-Laboratory/PCA-Seg}}\\
}

\begin{document}
\maketitle
\begin{abstract}
Recent advances in vision-language models (VLMs) have garnered substantial attention in open-vocabulary semantic and part segmentation (OSPS).  However, existing methods extract image-text alignment cues from cost volumes through a serial structure of spatial and class aggregations, leading to knowledge interference between class-level semantics and spatial context.  Therefore, this paper proposes a \textbf{simple} yet \textbf{effective} parallel cost aggregation (\textbf{PCA-Seg}) paradigm to alleviate the above challenge,  enabling the model to capture richer vision-language alignment information from cost volumes.   Specifically, we design an expert-driven perceptual learning (EPL) module that efficiently integrates semantic and contextual streams. It incorporates a multi-expert parser to extract complementary features from multiple perspectives. In addition, a coefficient mapper is designed to adaptively learn pixel-specific weights for each feature, enabling the integration of complementary knowledge into a unified and robust feature embedding.
Furthermore, we propose a feature orthogonalization decoupling (FOD) strategy to mitigate redundancy between the semantic and contextual streams, which allows the EPL module to learn diverse knowledge from orthogonalized features.
Extensive experiments on \textbf{eight} benchmarks show that each parallel block in PCA-Seg adds merely \textbf{0.35M} parameters while achieving state-of-the-art OSPS performance. 
\end{abstract}    

\section{Introduction}
Conventional segmentation \cite{mai2024rankmatch, chen2024knowledge,chen2024spatial, 11071876} is limited to a predefined set of categories, which restricts the model's ability to perceive the open-world visual environment. In recent years, vision-language foundation models (e.g., CLIP \cite{radford2021learning} and ALIGN \cite{jia2021scaling}), trained on large image-text paired datasets, have demonstrated remarkable open-vocabulary recognition capabilities.  Numerous studies \cite{zhou2022extract, yu2023convolutions, xu2023learning, Li_2025_CVPR} have adapted vision foundation models, initially designed for image-level recognition, to enable dense prediction tasks \cite{qu2025end,cai2024poly}. This adaptation has driven significant progress in open-vocabulary semantic and part segmentation (OSPS) \cite{wysoczanska2024clip, han2023open, ghiasi2022scaling, sun2023going} for arbitrary-class classification.

\begin{figure}[t]
	\centering
	\includegraphics[width=0.9\linewidth]{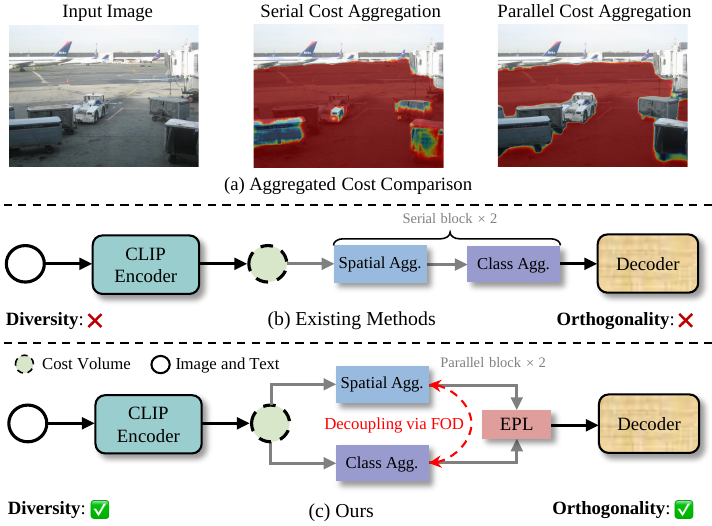}
    \vspace{-0.25cm}
	\caption{Motivation of  PCA-Seg. (a) Cost aggregation comparison among different methods for class \texttt{runway}. (b) The prevailing \textbf{\textit{serial}} architecture. (c) Our proposed \textbf{\textit{parallel}} cost aggregation paradigm. The FOD strategy decouples class-level semantics and spatial context, enabling the EPL module to parse diverse knowledge from the resulting orthogonalized features.}
	\label{fig1}
    \vspace{-0.2cm}
\end{figure}

Early OSPS methods \cite{liang2023open, ding2022decoupling, xu2022simple} freeze the parameters of CLIP within a two-stage framework to preserve its strong vision-language alignment. However, their significant computational overhead limits scalability and hinders further research. In contrast, current mainstream approaches \cite{wu2024clipself, Choi_2025_CVPR, cho2024cat, wang2025declip} adopt an end-to-end learning paradigm \cite{qu2024conditional,Cai_2025_ICCV,yin2025uncertainty}, enabling dense prediction through fine-tuning CLIP's attention layers. Representative open-vocabulary semantic segmentation (OVSS) methods, such as DeCLIP \cite{wang2025declip} and H-CLIP \cite{Peng_HCLIP_2025_CVPR}, along with open-vocabulary part segmentation (OVPS) approaches like PartCATSeg \cite{Choi_2025_CVPR}, follow a similar processing paradigm. These methods perform two consecutive operations  of spatial and class aggregation on the similarity matrix (cost volume) derived from visual and textual encoder features, as shown in Figure~\ref{fig1}(b).
Although the above design achieves excellent performance, its serial architecture induces \textit{knowledge interference between class-level semantics and spatial structural information}, thereby hindering the refinement of the cost volume and impairing image-text alignment. As illustrated in Figures~\ref{fig1}(a) and (b), applying spatial aggregation before class aggregation captures contextual structures but distorts the semantics of the \texttt{truck} category. The subsequent class aggregation further amplifies this semantic bias, causing the \texttt{truck} region to be misclassified as \texttt{runway}.

The aforementioned knowledge interference primarily arises from the cascading behavior of the serial structure, where the aggregation of one information source triggers a chain reaction in the subsequent aggregation of another.   This observation motivates the design of a parallel architecture that enables the two aggregation types to operate independently, thereby mitigating knowledge interference.
However, compared to the serial architecture, the baseline model, which captures both class semantics and spatial context through a single convolution, shows a slight decrease of 0.2\% mIoU on \texttt{PAS-20$^{b}$} (81.3\%→81.1\%). This raises a key challenge: \textit{How to efficiently integrate the two knowledge flows to facilitate diverse information learning?}

Therefore, this paper proposes a \textbf{\textit{simple}} yet \textbf{\textit{effective}} parallel cost aggregation (\textbf{PCA-Seg}) paradigm that alleviates the \textit{knowledge interference} between class-level semantics and spatial structural information, facilitating the extraction of richer image-text alignment cues from the cost volume, as shown in Figure~\ref{fig1}(c). Specifically, we design an expert-driven perceptual learning (EPL) module to efficiently integrate the two knowledge streams derived from class and spatial aggregation. The module comprises two key components: a multi-expert parser and a coefficient mapper. The multi-expert parser synthesizes both aggregation types to reveal complementary knowledge from multiple perspectives.
The coefficient mapper performs dimensionality reduction learning on semantic and spatial features, generating pixel-specific weight coefficients in a lightweight and adaptive manner. These coefficients are then used to emphasize the regions of the features generated by the multi-expert parser, ultimately producing more robust and unified embeddings. In contrast to the existing serial architecture, the EPL module fully exploits feature \textbf{diversity},  thereby activating more fine-grained information (see Figure~\ref{fig_redundancy_experts}).

\begin{figure}[t]
	\centering
	\includegraphics[width=0.98\linewidth]{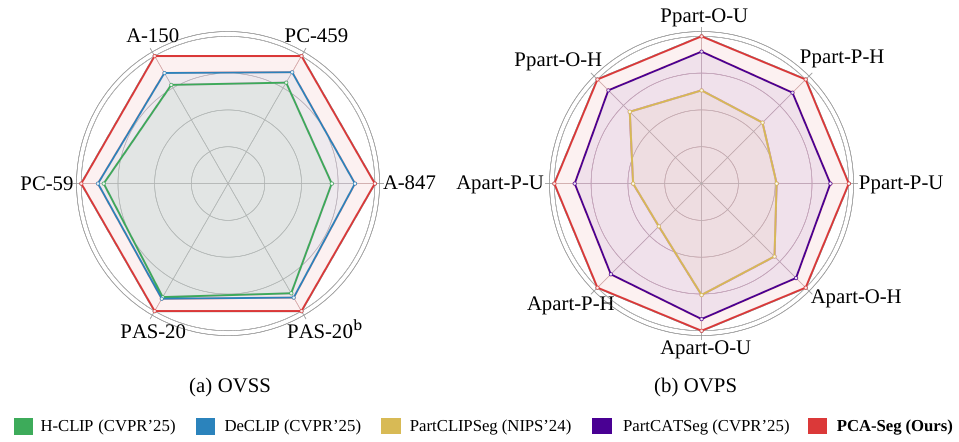}
    \vspace{-0.2cm}
	\caption{Comparison with other state-of-the-art methods on different benchmarks. ``Ppart'' and ``Apart'' are abbreviations for the Pascal-Part-116 \cite{wei2023ov} and ADE20K-Part-234 \cite{wei2023ov} datasets, respectively. `P' and `O' denote two different settings, namely Pred-All and Oracle-Obj. `H' represents the harmonic IoU over both seen and unseen classes, while `U' denotes the mIoU of unseen classes.}
	\label{fig_result}
    \vspace{-0.2cm}
\end{figure}

Class-level semantics and spatial structural information are typically regarded as knowledge representations along two distinct dimensions. These two knowledge flows should remain largely uncorrelated, encouraging the model to explore a broader feature space.
Motivated by this observation, we propose a feature orthogonalization decoupling (FOD) strategy that separates the representations generated by class and spatial aggregation through orthogonalization. This strategy ensures, from the knowledge source perspective, that the EPL module extracts more diverse and complementary knowledge, as depicted in Figure~\ref{fig1}(c). Concretely, since class-level semantic and spatial structural features remain correlated in the non-orthogonal state, an orthogonalization decoupling loss is designed to reduce their cosine similarity to zero, enforcing \textbf{orthogonality} between the two knowledge flows, as shown in Figure~\ref{fig_redundancy}(b). 
Ablation studies (see Table \ref{tab:component}) demonstrate that the FOD strategy enhances fine-grained feature learning within the EPL module, leading to a 0.9\% mIoU improvement  on \texttt{A-150}.


Compared with existing \textbf{\textit{serial}} blocks, each \textbf{\textit{parallel}} block in PCA-Seg adds only \textbf{0.35M} additional parameters and requires \textbf{0.96G} more GPU memory, while effectively mitigating \textit{knowledge interference} and achieving state-of-the-art performance on \textbf{eight} benchmarks,  which is illustrated in Figure~\ref{fig_result}.  To summarize, our contributions are as follows: (1) We revisit cost aggregation and propose PCA-Seg with a parallel architecture to mitigate knowledge interference between class-level semantics and spatial structural information, enabling robust image-text alignment learning from the cost volume. (2) Expert-driven perceptual learning is designed to perform multi-perspective, fine-grained analysis of knowledge obtained through class and spatial aggregation. (3) We further propose a feature orthogonalization decoupling strategy that  constrains class-level semantic and spatial structural features, allowing EPL to exploit richer knowledge. (4) We conduct extensive experiments on eight benchmarks, and the results demonstrate that the proposed method achieves state-of-the-art performance.

\section{Related Work}
\label{sec:related_work}

\noindent \textbf{Open-Vocabulary Semantic and Part Segmentation.} Recently, the emergence of vision-language models (VLMs) \cite{radford2021learning,li2021align,jia2021scaling,shen2025fine} has substantially advanced the development of open-vocabulary segmentation \cite{qin2023freeseg,NEURIPS2024_11195878}, enabling recognition across arbitrary categories. For open-vocabulary semantic segmentation (OVSS), early studies \cite{ding2022decoupling, ghiasi2022scaling, xu2023open, liang2023open, han2023open} commonly employ a two-stage training framework, where a robust class-agnostic mask generator \cite{cheng2022masked} is first trained to delineate object regions, and the resulting masks are then fed into CLIP \cite{radford2021learning} for classification. The performance of two-stage methods is highly dependent on the quality of the generated masks. 
To address this limitation, numerous end-to-end approaches \cite{xu2023side,xie2024sed,shan2024open,wang2025declip} have been proposed to fine-tune specific CLIP parameters, such as attention layers, thereby enabling dense prediction and precise localization. Notably, H-CLIP \cite{Peng_HCLIP_2025_CVPR} and HyperCLIP \cite{Peng_2025_CVPR} focus on designing novel mechanisms to fine-tune CLIP parameters within hyperbolic space.  To achieve a more fine-grained understanding of object components, early part segmentation methods \cite{choudhury2021unsupervised,he2023compositor,van2023pdisconet} employ contrastive and self-supervised learning techniques to accurately delineate target part regions. Recent advances in open-vocabulary part segmentation (OVPS), including VLPart \cite{sun2023going}, OV-PARTS \cite{wei2023ov}, PartGLEE \cite{li2024partglee}, and PartCLIPSeg \cite{NEURIPS2024_f7f47a73}, build upon the OVSS framework and achieve superior performance by integrating prior knowledge, designing effective prompt strategies, and leveraging object-level contextual information.

\noindent \textbf{Open-Vocabulary Cost Aggregation.} Reducing matching errors and improving generalization in dense visual  correspondence \cite{hong2024unifying,liu2020semantic,min2019hyperpixel,cai2025beyond}  has been effectively achieved through cost aggregation \cite{chen2023costformer,Choi_2025_CVPR,cho2024cat,hong2022cost,truong2020glu}, as evidenced by the work of CATs \cite{cho2021cats} and CATs++ \cite{cho2022cats++} within Transformer-based architectures. Building on this, prominent methods such as CAT-Seg \cite{cho2024cat}, H-CLIP \cite{Peng_HCLIP_2025_CVPR}, and DeCLIP \cite{wang2025declip} incorporate cost aggregation into  OVSS. They process the cost volume through two consecutive blocks of spatial and class aggregation, achieving impressive performance.   Similarly, PartCATSeg \cite{Choi_2025_CVPR} incorporates cost aggregation into OVPS, laying a solid foundation for further research in this area. \textit{In contrast to the aforementioned \textbf{serial} aggregation methods, PCA-Seg employs \textbf{parallel} cost aggregation to decouple knowledge, allowing the model to learn richer image–text alignment information from the cost volume.}

\begin{figure*}[t]
	\centering
	\includegraphics[width=1\linewidth]{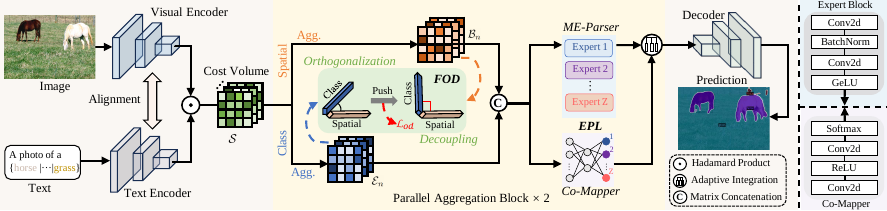}
    \vspace{-0.6cm}
	\caption{The framework of PCA-Seg on OVSS. Image and text features from CLIP’s visual and text encoders are combined via the Hadamard product to construct the cost volume $\Scal$. This volume is then processed simultaneously through spatial and class aggregation to produce the spatial context feature $\Bcal_{n}$ and the class-level semantic representation $\Ecal_{n}$. The multi-expert (ME-) parser extracts diverse and complementary knowledge from these two streams across multiple perspectives, while the coefficient (Co-) mapper adaptively learns weights to integrate the features parsed by the experts. $\Bcal_{n}$ and $\Ecal_{n}$ are decoupled using the feature orthogonalization decoupling (FOD) strategy to reduce redundancy, thereby providing enriched  representations for expert-driven perceptual learning (EPL).}
	\label{fig3_net}
    \vspace{-0.1cm}
\end{figure*}

\section{Method}
\label{sec:method}
OVSS and OVPS primarily differ in terms of class granularity. For clarity, this section provides a detailed description of the proposed PCA-Seg method, illustrating its framework using OVSS, as shown in Figure~\ref{fig3_net}. We begin with an overview of the preliminary work ($\S\ref{preliminary}$), followed by the design of expert-driven perceptual learning ($\S\ref{PSI}$) and feature orthogonalization decoupling ($\S\ref{FOP}$). Finally, the overall training objective ($\S\ref{OTO}$) is formulated.

\subsection{Preliminary}
\label{preliminary}
Models for open-vocabulary semantic and part segmentation are trained on the seen category set $\Ccal^t$ and evaluated on a validation set $\Ccal^v$, which may include both seen categories from $\Ccal^t$ and unseen categories. By leveraging the strong image-text alignment of CLIP \cite{radford2021learning}, the training process facilitates generalization to unseen categories without additional supervision and enables the assessment of knowledge transfer from seen to unseen concepts. The following section describes the cost volume generation and serial architecture within the methods CAT-Seg \cite{cho2024cat}, DeCLIP \cite{wang2025declip}, and PartCATSeg \cite{Choi_2025_CVPR}.

\noindent\textbf{Cost Volume Generation.} Given an input image $\Ical$ and its corresponding text description $\Tcal$, CLIP generates the visual embedding $\Fcal^v \in \mathbb{R}^{H \times W \times C}$ and the textual embedding $\Fcal^t \in \mathbb{R}^{N \times C}$ through its visual and text encoders. Here, $H$ and $W$ denote the height and width, respectively, $N$ represents the total number of categories, and $C$ indicates the feature dimension. Following CAT-Seg \cite{cho2024cat}, the final attention layer of the visual encoder is replaced with two $1 \times 1$ convolutional layers to improve object localization accuracy. The cost volume $\Scal \in \mathbb{R}^{H \times W \times N}$ is then generated according to the following formulation:
\begin{equation}
    \label{eq1}
    \Scal(i,j)=\frac{\Fcal^{v}(i)\cdot \Fcal^{t}(j)}{||\Fcal^{v}(i)||\cdot||\Fcal^{t}(j)||}. 
\end{equation}
$||\cdot||$ denotes the L2 norm of the feature. $\Scal(i,j)$ represents the cosine similarity between the $i$-th pixel of the visual embedding and the $j$-th category of the text embedding.  $\Scal$ is then convolved with a $1 \times 1$ convolution to produce the cost volume embedding $\Vcal_{1} \in \mathbb{R}^{H \times W \times N \times C}$.

\noindent\textbf{Serial Architecture.}  The cost volume embedding is refined through two consecutive serial blocks of spatial and class aggregation, enhancing image-text alignment. Specifically, the embedding $\Vcal_{1}$ is first processed by spatial aggregation $\Phi_{n} (\cdot)$ to capture structural context across spatial dimensions. It is then refined through class aggregation $\Gamma_{n} (\cdot)$ to exploit category-level semantic features, formulated as follows:
\vspace{-2pt}
\begin{equation}
    \label{eq2}
     \Vcal_{n+1}  =  \textcolor{red}{\Gamma_{n} (}\textcolor{brown}{\Phi_{n} (\Vcal_{n})}\textcolor{red}{)}.\\
     \vspace{-2pt}
\end{equation}
Here, $n$ is the index, taking values in $\{1,2\}$. Both $\Vcal_n$ and $\Vcal_{n+1}$ maintain the same size throughout. Notably, spatial aggregation utilizes the Swin Transformer \cite{liu2021swin}, while class aggregation employs a linear variant \cite{katharopoulos2020transformers}.

\noindent\textbf{Key Motivation.} Although the serial architecture achieves notable performance, its inherent cascading effect entangles class and spatial aggregation.  This causes biases from the preceding step to propagate into subsequent information learning, resulting in knowledge interference, as shown in Figure \ref{fig1} (a).  This motivates the design of a \textit{\textbf{parallel}} aggregation architecture to decouple the learning of class-level semantics and spatial contextual information, thereby alleviating the aforementioned issue. The key challenge lies in effectively aggregating these two types of knowledge for more robust and fine-grained image-text alignment, which is detailed in the following sections.


\subsection{Expert-driven Perceptual Learning}
\label{PSI}
The two knowledge streams in the \textbf{\textit{parallel}} architecture are designed to complement each other, thereby enhancing the image-text alignment capability of the model. To this end, we propose the expert-driven perceptual learning (EPL) module, which efficiently integrates the two streams. In EPL, a multi-expert  parser disentangles and refines class-level semantics and spatial contextual knowledge, extracting diverse and complementary information. These features are then adaptively fused using coefficients learned by a coefficient mapper, resulting in unified and robust feature embeddings.

\noindent \textbf{Parallel Architecture.}  Unlike the sequential aggregation in a serial manner, the cost volume embedding $\Vcal_{1}$ is processed simultaneously by spatial and class-level  aggregation to produce the corresponding spatial context feature $\Bcal_{n}$ and class-level semantic representation $\Ecal_{n}$:
\vspace{-3pt}
\begin{equation}
    \label{eq3}
    \Bcal_{n}=\textcolor{brown}{\Phi_{n} (\Vcal_{n} )}, \ \ \ \ \  \Ecal_{n} = \textcolor{red}{\Gamma_{n} (\Vcal_{n})}.
    \vspace{-3pt}
\end{equation}
The above equation enables the decoupled learning of the two knowledge streams, with $n$ defined consistently as in Equation \ref{eq2}. Each pair 
\{$\Phi_{n}(\cdot)$, $\Gamma_{n}(\cdot)$\} represents the $n$-th parallel block.

\noindent\textbf{Multi-Expert Parser.} We first concatenate $\Bcal_{n}$ and $\Ecal_{n}$ along the channel dimension to obtain the feature $\Acal$, which is subsequently processed by multiple expert blocks $\{\Upsilon_{z}(\cdot)\}_{z=1}^{\Zcal}$ to extract diverse and complementary  knowledge representations $\{\Dcal_{z}\}_{z=1}^{\Zcal}$ from different perspectives, as shown below: 
\vspace{-3pt}
\begin{equation}
    \label{eq4}
    \Dcal_{z}= \Upsilon_{z}(\Acal).
    \vspace{-3pt}
\end{equation}
Each expert block adopts the same structure, consisting of two convolution operations with a BatchNorm layer in between, followed by a GeLU activation for the output, introducing only the additional parameters of \textbf{8.5$\times$10$^{\textbf{-2}}$ M}.
Ablation results indicate that the model achieves the best performance when the number of expert blocks $\Zcal$ is set to 4.  The feature redundancy quantified by CCA \cite{raghu2017svcca} (canonical correlation analysis), shown in Figure \ref{fig_redundancy_experts} of the experiment section, highlights that each expert learns complementary and diverse knowledge.

\noindent \textbf{Coefficient Mapper.} The next step is to fuse the outputs  $\{\Dcal_{z}\}_{z=1}^{\Zcal}$ from the multi-expert parser to obtain  fine-grained and robust feature embeddings.  The simplest approach involves using a single convolution for fusion, but the experimental results show only a marginal improvement in performance, as shown in Table \ref{tab:fusion}. This outcome is attributed to the convolution disrupting complementary knowledge from multiple perspectives, thereby hindering the efficient integration of diverse information. Another aggregation strategy is weighted aggregation. Unlike previous approaches (e.g., average) that apply fixed weights for feature aggregation, the proposed coefficient mapper assigns pixel-specific weights. It allows the model to autonomously learn and assign pixel-level weights to the features parsed by each expert in a lightweight and adaptive manner, ensuring efficient integration of diverse knowledge while preserving complementary information.

Feeding the feature $\Acal$ into the coefficient mapper $\Theta(\cdot)$  produces the  coefficients $\{\Pcal_{z}\}_{z=1}^{\Zcal}$:
\vspace{-3pt}
\begin{equation}
    \label{eq5}
    \Pcal_1,\Pcal_2,\cdot\cdot \cdot, \Pcal_\Zcal =\Theta (\Acal).
    \vspace{-3pt}
\end{equation}
The coefficient mapper consists of two convolutional layers with a ReLU activation in between, followed by a softmax function to ensure the coefficients sum to 1. It is lightweight and capable of dynamic weight learning, containing only \textbf{8$\times$10$^{\textbf{-3}}$ M} parameters.

\noindent  \textbf{Efficient Integration.} Each coefficient in  $\{\Pcal_{z}\}_{z=1}^{\Zcal}$ is paired with an element in $\{\Dcal_{z}\}_{z=1}^{\Zcal}$ through indexing, and their element-wise products are used to aggregate the knowledge features parsed by each expert, generating high-quality and robust feature embeddings $\Rcal$:
\vspace{-3pt}
\begin{equation}
    \label{eq6}
    \Rcal = \sum_{z=1}^{\Zcal}\Pcal_z \cdot \Dcal_z.
    \vspace{-3pt}
\end{equation}
The proposed expert-driven perceptual learning introduces only \textbf{0.35M} parameters, yet achieves a 0.9\% mIoU gain on the \texttt{PAS-20$^{b}$} dataset (Table~\ref{tab:component}), with consistent improvements observed in part segmentation. These results highlight its efficiency and effectiveness.

\subsection{Feature Orthogonalization Decoupling}
\label{FOP}
Class-level semantics and spatial context knowledge represent information across two distinct dimensions. Ideally, the relationship between these two knowledge streams should remain independent, which is a key factor in the success of the expert-driven perceptual learning module. More specifically, the two unrelated feature representations ensure that EPL can learn the diversity and complementary information essential for segmentation from a broad feature space.   Therefore, we further propose a feature orthogonalization decoupling strategy to constrain the relationship between class-level semantics and spatial context knowledge, thereby promoting orthogonality.

The correlation between the two-stream feature knowledge, $\Bcal_{n}$ and $\Ecal_{n}$, is measured using cosine similarity:
\vspace{-3pt}
\begin{equation}
    \label{eq7}
    \Mcal_{i}=\frac{\Bcal_{n}(i)\cdot \Ecal_{n}(i)}{||\Bcal_{n}(i)||\cdot||\Ecal_{n}(i)||}.
    \vspace{-3pt}
\end{equation}
Here, $\Mcal_{i}$ represents the similarity between the features of the $i$-th pixel at the same position in both representations. Minimizing this similarity reduces the correlation between the two streams of knowledge. However, the presence of negative values in cosine similarity causes gradient backpropagation to become imprecise, steering the optimization in the wrong direction. To address this issue, we propose the following feature orthogonalization decoupling loss:
\vspace{-3pt}
\begin{equation}
    \label{eq8}
    \Lcal_{od} = \frac{1}{H \times W} \sum_{i=1}^{H \times W}M_{i}^{2}.
    \vspace{-3pt}
\end{equation}
Squaring is employed to ensure the loss remains positive and converges to zero, thus facilitating the orthogonalization of both streams.

\subsection{Overall Training Objective}
\label{OTO}
The robust unified feature embeddings $\Rcal$ are passed through the decoder $\Gcal(\cdot)$ to generate the predicted results, which incur the following supervised loss relative to the ground truth $\Ycal$:
\vspace{-3pt}
\begin{equation}
    \label{eq9}
    \Lcal_{sup} = \frac{1}{H \times W} \sum_{i=1}^{H \times W}{{\ell _{ce}}}(\Gcal(\Rcal),\Ycal).
    \vspace{-3pt}
\end{equation}
Total loss $\Lcal$ for network parameter optimization is as follows:
\vspace{-3pt}
\begin{equation}
    \label{eq10}
    \Lcal = \Lcal_{sup} + \lambda \Lcal_{od}.
    \vspace{-3pt}
\end{equation}
Here, $\lambda$ represents the weight of the feature orthogonalization decoupling loss $\Lcal_{od}$. Extensive ablation studies show that the optimal value for $\lambda$ is 0.01. Consistent with prior research \cite{cho2024cat, wang2025declip, Choi_2025_CVPR}, we update only the query and value within the CLIP encoder, while optimizing the parameters outside the encoder to minimize computational overhead.
\section{Experiments}
\label{experiments}

\begin{table*}[!t]
    \begin{center}
    \renewcommand\arraystretch{1}
    \caption{Comparison of performance  with other state-of-the-art methods  on multiple standard benchmarks in open-vocabulary semantic segmentation. See $\S\ref{Compare_method}$ for details.}
    \vspace{-0.8em}
    \resizebox{1.0\textwidth}{!}{
    {\scriptsize
    \begin{tabular}{r||ccc||cccccc}
    \spthickhline
       \rowcolor[rgb]{0.92,0.92,0.92} Method &  Publication & Backbone  & Training Dataset  & \texttt{A-847} & \texttt{PC-459} & \texttt{A-150} & \texttt{PC-59} & \texttt{PAS-20} & \texttt{PAS-20$^{b}$} \\
       \hline \hline
        LSeg+~\cite{ghiasi2022scaling}& ECCV'22 & ALIGN  & COCO-Stuff   & 2.5 & 5.2 & 13.0 & 36.0 & - & 59.0 \\
        \rowcolor{table_content}ZegFormer~\cite{ding2022decoupling} & CVPR'22 &  ViT-B/16  & COCO-Stuff-156    & 4.9 & 9.1 & 16.9 & 42.8 & 86.2 & 62.7 \\
        ZSseg~\cite{xu2022simple} & ECCV'22 & ViT-B/16  & COCO-Stuff    & 7.0 & - & 20.5 & 47.7 & 88.4 & - \\
        \rowcolor{table_content}OVSeg~\cite{liang2023open} & CVPR'23 & ViT-B/16  & COCO-Stuff + Caption   &  7.1 & 11.0 & 24.8 & 53.3 & 92.6 & - \\
        GKC~\cite{han2023global} & ICCV'23 &   ViT-B/16  & COCO Panoptic   & 3.5 & 7.1 & 18.8 & 45.2 &  83.2 & - \\
        \rowcolor{table_content} EBSeg \cite{shan2024open} & CVPR'24 & ViT-B/16 & COCO-Stuff & 11.1 & 17.3 & 30.0 & 56.7 & 94.6 & - \\ 
        CAT-Seg \cite{cho2024cat}   & CVPR'24 & ViT-B/16 & COCO-Stuff & 12.0 & 19.0 & 31.8 & 57.5 & 94.6 & 77.3  \\
        \rowcolor{table_content}SED \cite{xie2024sed} & CVPR'24 &  ConvNeXt-B & COCO-Stuff & 11.4 & 18.6 & 31.6 & 57.3 & 94.4 & -  \\
        Mask-Adapter \cite{Li_2025_CVPR} & CVPR'25  &  ConvNeXt-B  & COCO-Stuff & 14.2 & 17.9 & 35.6 & 58.4 & 95.1 & -  \\
        \rowcolor{table_content} DeCLIP \cite{wang2025declip} & CVPR'25  & ViT-B/16  & COCO-Stuff & 15.3 & 21.4 & 36.3 & 60.6 & 96.6 & 81.3  \\
        H-CLIP \cite{Peng_HCLIP_2025_CVPR} & CVPR'25 & ViT-B/16 & COCO-Stuff & 12.5 & 19.4 & 32.4 & 57.9 & 95.2 & 78.2 \\
        \hdashline
        \rowcolor{unitcolor3} \textbf{PCA-Seg (Ours)} & - & ViT-B/16  & COCO-Stuff  & \textbf{16.1} & \textbf{22.3}
        & \textbf{38.1} &  \textbf{62.2}&  \textbf{97.3} & \textbf{82.7} \\
        
        \hline \hline
        OVSeg~\citep{liang2023open} & CVPR'23 &  ViT-L/14  & COCO-Stuff + Caption  & 9.0 & 12.4 & 29.6 & 55.7 & 94.5 & - \\
        \rowcolor{table_content} ODISE~\citep{xu2023open} & CVPR'23 & ViT-L/14  & COCO-Panoptic   & 11.1 & 14.5 & 29.9 & 57.3 & - & 84.6 \\
        EBSeg \cite{shan2024open} & CVPR'24 & ViT-L/14 & COCO-Stuff & 13.7 & 21.0 & 32.8 & 60.2 & 96.4 & -\\ 
        \rowcolor{table_content}CAT-Seg \cite{cho2024cat}   & CVPR'24 & ViT-L/14 & COCO-Stuff & 16.0 & 23.8 & 37.9 & 63.3 & 97.0 & 82.5 \\
        SED \cite{xie2024sed} & CVPR'24 &  ConvNeXt-L  & COCO-Stuff & 13.9 & 22.6 & 35.2 & 60.6 & 96.1 & - \\
        \rowcolor{table_content}Mask-Adapter \cite{Li_2025_CVPR} & CVPR'25 &  ConvNeXt-L  & COCO-Stuff & 16.2 & 22.7 & 38.2 & 60.4 & 95.8 & - \\
        DeCLIP \cite{wang2025declip} & CVPR'25 & ViT-L/14  & COCO-Stuff & 17.6 & 25.9 & 40.7 & 63.9 & 97.7 & 83.9 \\
        \rowcolor{table_content}H-CLIP \cite{Peng_HCLIP_2025_CVPR} & CVPR'25 & ViT-L/14 & COCO-Stuff & 16.5 & 24.2 & 38.4 & 64.1 & 97.7 & 83.2\\
        LLMFormer \cite{shi2025llmformer} & IJCV'25 & ViT-L/14 & COCO-Stuff & 16.5 & 25.4 & 38.5 & 64.2 & 96.8 & - \\
        \hdashline
         \rowcolor{unitcolor3} \textbf{PCA-Seg (Ours)} & - &  ViT-L/14  & COCO-Stuff  & \textbf{18.3}  & \textbf{26.7} & \textbf{41.5} & \textbf{64.7}  & \textbf{98.4} & \textbf{84.4} \\

        \hline
    \end{tabular}
    }
    }
    \label{tab:semantic_results}
    \vspace{-1.1em}
    \end{center}
\end{table*}

\subsection{Experimental Setup}
\textbf{Datasets.} For OVSS, our model is trained on the COCO-Stuff \cite{caesar2018coco} dataset, which contains 118k high-quality annotations across 171 categories, and directly evaluated on multiple standard benchmarks including ADE20K \cite{zhou2019semantic},  PASCAL-Context \cite{mottaghi2014role}, and PASCAL VOC \cite{everingham2010pascal} to comprehensively  assess  performance. These datasets are organized by class into \texttt{A-847}, \texttt{A-150}, \texttt{PC-459}, \texttt{PC-59}, \texttt{PAS-20}, and \texttt{PAS-20$^{b}$}. For OVPS, we evaluate PCA-Seg on Pascal-Part-116 \cite{chen2014detect,wei2023ov} and ADE20K-Part-234 \cite{zhou2017scene,wei2023ov}.

\noindent \textbf{Evaluation Metrics.} We evaluate the model on OVSS using the mean Intersection-over-Union (mIoU) metric. For OVPS, following prior works \cite{Choi_2025_CVPR,li2024partglee}, we adopt two evaluation protocols to assess zero-shot generalization. The first protocol, \textbf{Pred-All}, introduced in PartCLIPSeg \cite{NEURIPS2024_f7f47a73}, relies solely on the model’s predictions. The second protocol, \textbf{Oracle-Obj}, follows the OV-PARTS \cite{wei2023ov} setting and provides ground-truth masks and labels to guide segmentation. In addition to mIoU, we evaluate generalization using the harmonic mean (h-IoU) between seen and unseen classes.

\noindent \textbf{Implementation Details.} The PCA-Seg model is implemented in PyTorch using the Detectron2 framework, and all experiments are conducted on eight NVIDIA A6000 GPUs. For OVSS, both ViT-B/16 and ViT-L/14 backbones are employed, with a learning rate of $3\times10^{-4}$ and 80k training iterations on the COCO-Stuff dataset. For OVPS, ViT-B/16 is used as the backbone. The learning rates for Pascal-Part-116 and ADE20K-Part-234 are set to $2\times10^{-4}$ and $4\times10^{-4}$, with 30k and 20k training iterations, respectively. The batch size is set to 8, and AdamW \cite{loshchilov2017decoupled} is adopted for optimization. 
Image resolutions follow prior works \cite{wang2025declip,Choi_2025_CVPR}.

\begin{table*}[ht]
    \centering
    \renewcommand\arraystretch{1}
    \caption{Comparison of zero-shot performance with other state-of-the-art methods  on the Pascal-Part-116 and ADE20K-Part-234 datasets in open-vocabulary part segmentation. See $\S\ref{Compare_method}$ for details.}
    \vspace{-0.8em}
    \small
    \resizebox{\linewidth}{!}{
        \begin{tabular}{r||cc||ccc|ccc|ccc|ccc}
    \spthickhline
    \rowcolor[rgb]{0.92,0.92,0.92} 
    & & & \multicolumn{6}{c|}{Pascal-Part-116} & \multicolumn{6}{c}{ADE20K-Part-234} \\  
    \rowcolor[rgb]{0.92,0.92,0.92} 
    & & & \multicolumn{3}{c|}{Pred-All} & \multicolumn{3}{c|}{Oracle-Obj} & \multicolumn{3}{c|}{Pred-All} & \multicolumn{3}{c}{Oracle-Obj} \\  
    \rowcolor[rgb]{0.92,0.92,0.92} 
    \multicolumn{1}{r||}{\multirow{-3}{*}{Method}} & 
    \multicolumn{1}{c}{\multirow{-3}{*}{Publication}} & 
    \multicolumn{1}{c||}{\multirow{-3}{*}{Backbone}} &
    Seen & Unseen & h-IoU & Seen & Unseen & h-IoU &
    Seen & Unseen & h-IoU & Seen & Unseen & h-IoU \\ 
    \hline \hline
    ZSSeg+ \cite{xu2022simple} & ECCV'22 & ResNet-50 & 38.1 & 3.4  & 6.2 & 54.4 & 19.0 & 28.2 & 32.3 & 0.9 & 1.7 & 43.2 & 27.8 & 33.9 \\
    \rowcolor{table_content} CLIPSeg \cite{luddecke2022image,wei2023ov}   & CVPR'22 & ViT-B/16 & 27.8 & 13.3 & 18.0 & 48.9 & 27.5 & 35.2 & 3.1 & 0.6 & 0.9 & 38.2 & 30.9 & 34.2\\
    VLPart \cite{sun2023going} & ICCV'23 & ResNet-50 & 35.2 & 9.0 & 14.4 & 42.6 & 18.7 & 26.0 & - & - & - & - & -& - \\
    \rowcolor{table_content} CAT-Seg \cite{cho2024cat,wei2023ov} & CVPR'24 & ViT-B/16 & 36.8 &  23.4 & 28.6 & 43.8 & 27.7 & 33.9 & 7.0 & 2.4 & 3.5 & 33.8 & 25.9 & 29.3\\
    PartGLEE \cite{li2024partglee} & ECCV'24 & ResNet-50 & - & - & - & 57.4 & 27.4 & 37.1 & - & - &- & 51.3 & 35.3 & 41.8 \\
    \rowcolor{table_content} PartCLIPSeg \cite{NEURIPS2024_f7f47a73} & NeurIPS'24& ViT-B/16 & 43.9 & 23.6 & 30.7 & 50.0 & 31.7 & 38.8 & 14.2 & 9.5 & 11.4 & 38.4 & 38.8 & 38.6 \\
    PartCATSeg \cite{Choi_2025_CVPR} & CVPR'25 & ViT-B/16 & 52.6 & 40.5 & 45.8 & 57.5 & 44.9 & 50.4 & 38.9 & 17.6 & 24.2 & 53.1 & 47.2 & 50.0\\
    \hdashline
    \rowcolor{unitcolor3} \textbf{PCA-Seg (Ours)} & - & ViT-B/16 & \textbf{57.2} & \textbf{43.5} & \textbf{49.3} & \textbf{60.5} & \textbf{47.1} & \textbf{52.9} & \textbf{40.8} & \textbf{19.2} & \textbf{26.1} & \textbf{56.3} &  \textbf{48.2} & \textbf{51.8} \\ \hline
        \end{tabular}}
    \label{tab:part_results}
    \vspace{-0.6em}
\end{table*}

\subsection{Comparison with State-of-the-Art Methods}
\label{Compare_method}

\noindent \textbf{Open-Vocabulary Semantic Segmentation.} Table \ref{tab:semantic_results} presents a performance comparison between PCA-Seg and other state-of-the-art methods across six standard OVSS benchmarks. The experimental results demonstrate that our method consistently outperforms others, regardless of whether the ViT-B/16 or ViT-L/14 backbone is used. Specifically, with ViT-B/16 as the backbone, PCA-Seg surpasses H-CLIP \cite{Peng_HCLIP_2025_CVPR} by 3.6\%, 2.9\%, 5.7\%, 4.3\%, 2.1\%, and 4.5\% mIoU on the \texttt{A-847}, \texttt{PC-459}, \texttt{A-150}, \texttt{PC-59}, \texttt{PAS-20}, and \texttt{PAS-20$^{b}$} datasets, respectively. When compared to DeCLIP \cite{wang2025declip}, which uses a \textbf{\textit{serial}} aggregation architecture, our method achieves mIoU improvements of 0.8\%, 0.9\%, 1.8\%, 1.6\%, 0.7\%, and 1.4\%. A similar performance advantage is consistently observed with the ViT-L/14 backbone. These experimental results highlight the remarkable effectiveness of PCA-Seg, achieving new state-of-the-art performance on OVSS.

\noindent \textbf{Open-Vocabulary Part Segmentation.} Table \ref{tab:part_results} summarizes the performance of our method in comparison with other state-of-the-art approaches on two OVPS benchmarks.  Across both the Pascal-Part-116 and ADE20K-Part-234 datasets, the proposed method demonstrates superior performance. For instance, on Pascal-Part-116, our approach surpasses PartCATSeg \cite{Choi_2025_CVPR}, which employs a \textbf{\textit{serial}} aggregation architecture, by 3.5\% and 2.5\% in h-IoU under the Pred-All and Oracle-Obj settings, respectively. Similarly, on ADE20K-Part-234, it outperforms PartCATSeg \cite{Choi_2025_CVPR} by 1.9\% and 1.8\%, respectively. It is evident that PCA-Seg continues to achieve SOTA performance on OVPS.

\begin{figure*}[!t]
    \centering
    \begin{subfigure}[t]{0.1206\textwidth} \includegraphics[height=1.9cm,width=\textwidth]{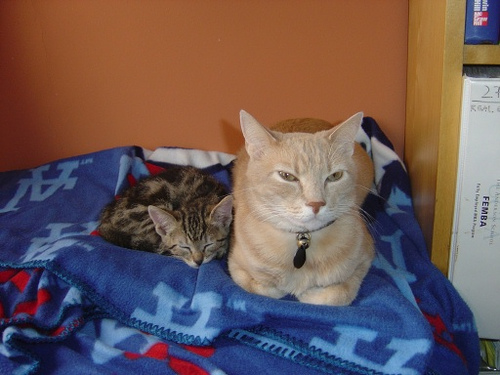} \end{subfigure}
    \begin{subfigure}[t]{0.1206\textwidth} \includegraphics[height=1.9cm,width=\textwidth]{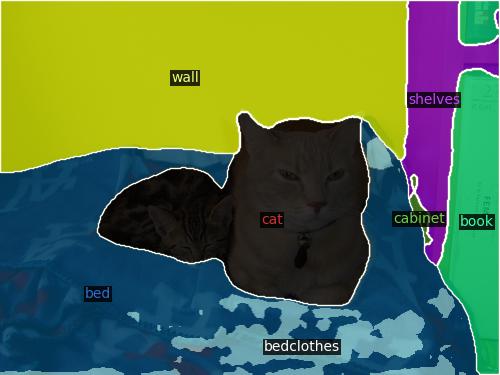} \end{subfigure}
    \begin{subfigure}[t]{0.1206\textwidth} \includegraphics[height=1.9cm,width=\textwidth]{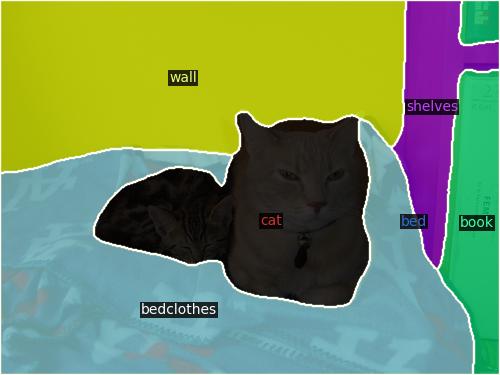} \end{subfigure}
    \begin{subfigure}[t]{0.1206\textwidth} \includegraphics[height=1.9cm,width=\textwidth]{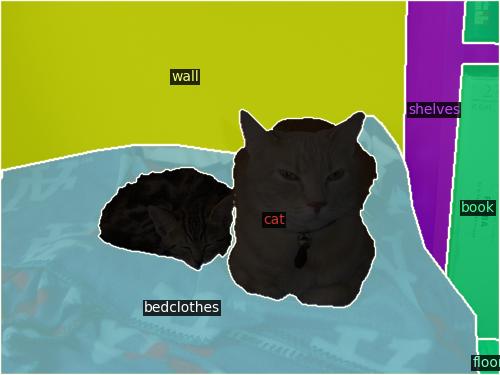} \end{subfigure}
    \begin{subfigure}[t]{0.1206\textwidth} \includegraphics[height=1.9cm,width=\textwidth]{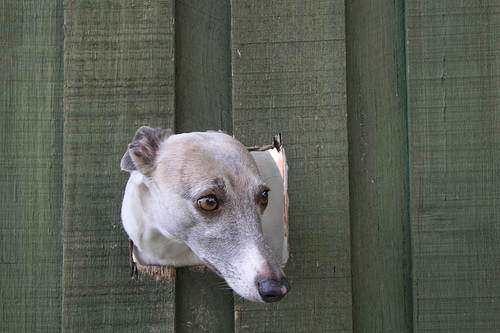} \end{subfigure}
    \begin{subfigure}[t]{0.1206\textwidth} \includegraphics[height=1.9cm,width=\textwidth]{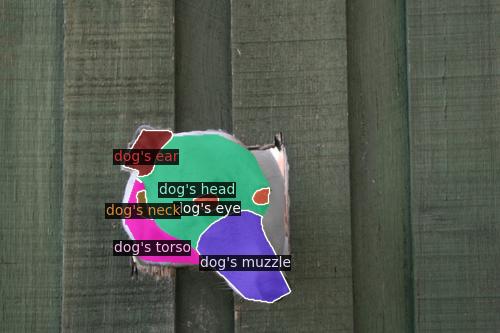} \end{subfigure}
    \begin{subfigure}[t]{0.1206\textwidth} \includegraphics[height=1.9cm,width=\textwidth]{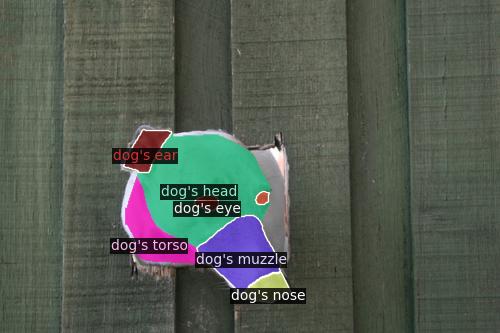} \end{subfigure}
    \begin{subfigure}[t]{0.1206\textwidth} \includegraphics[height=1.9cm,width=\textwidth]{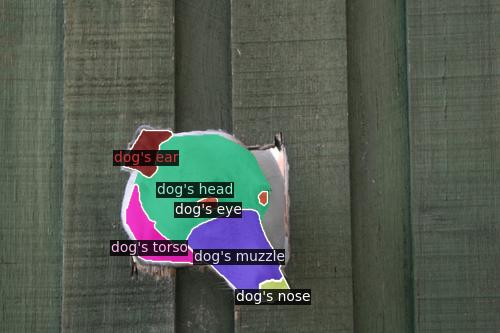} \end{subfigure}
    \begin{subfigure}[t]{0.1206\textwidth} \includegraphics[height=1.9cm,width=\textwidth]{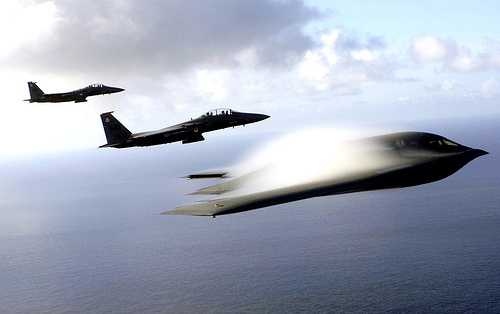} \end{subfigure}
    \begin{subfigure}[t]{0.1206\textwidth} \includegraphics[height=1.9cm,width=\textwidth]{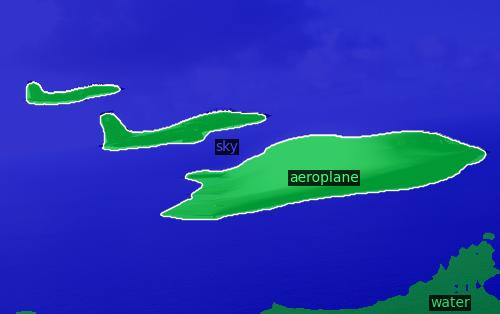} \end{subfigure}
    \begin{subfigure}[t]{0.1206\textwidth} \includegraphics[height=1.9cm,width=\textwidth]{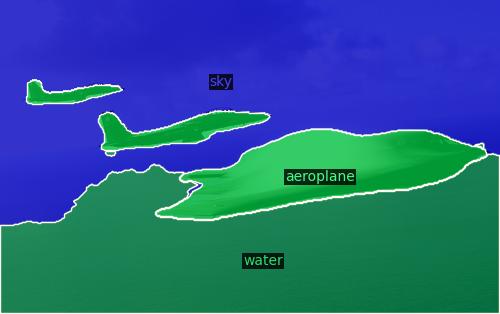} \end{subfigure}
    \begin{subfigure}[t]{0.1206\textwidth} \includegraphics[height=1.9cm,width=\textwidth]{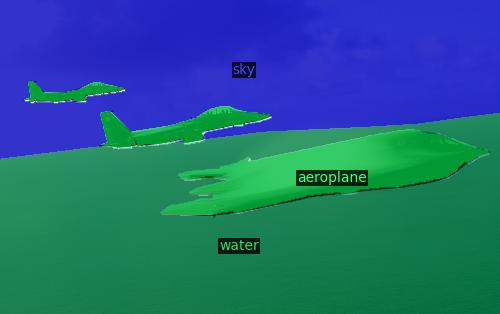} \end{subfigure}
    \begin{subfigure}[t]{0.1206\textwidth}
\includegraphics[height=1.9cm,width=\textwidth]{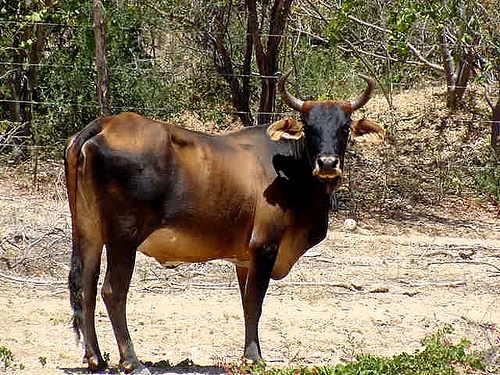}
    \end{subfigure}
    \begin{subfigure}[t]{0.1206\textwidth}
\includegraphics[height=1.9cm,width=\textwidth]{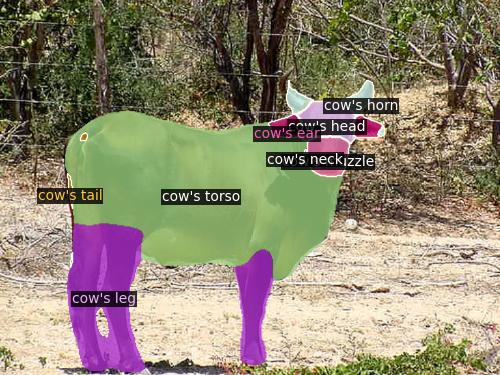}
    \end{subfigure}
    \begin{subfigure}[t]{0.1206\textwidth}
\includegraphics[height=1.9cm,width=\textwidth]{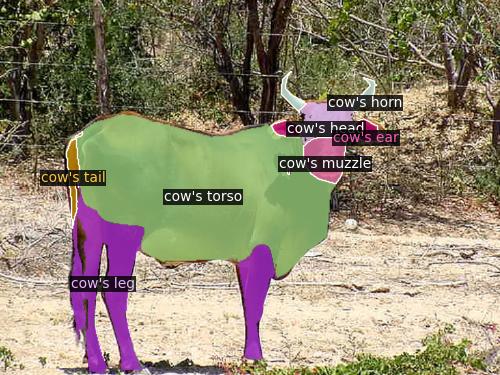}
    \end{subfigure}
    \begin{subfigure}[t]{0.1206\textwidth}
\includegraphics[height=1.9cm,width=\textwidth]{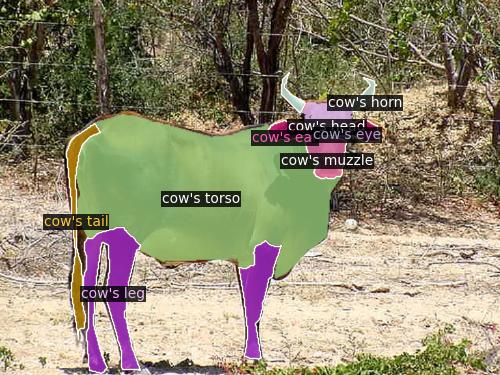}
    \end{subfigure}
    \begin{subfigure}[t]{0.1206\textwidth}
\includegraphics[height=1.9cm,width=\textwidth]{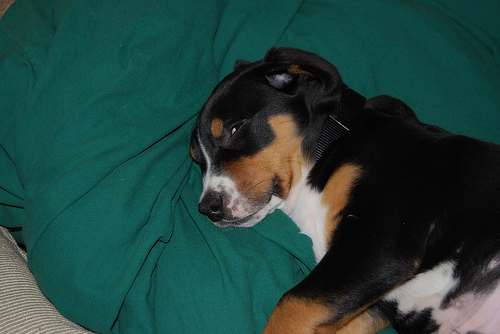}
    \end{subfigure}
    \begin{subfigure}[t]{0.1206\textwidth}
\includegraphics[height=1.9cm,width=\textwidth]{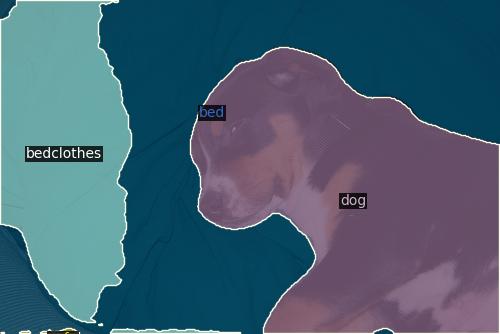}
    \end{subfigure}
    \begin{subfigure}[t]{0.1206\textwidth} \includegraphics[height=1.9cm,width=\textwidth]{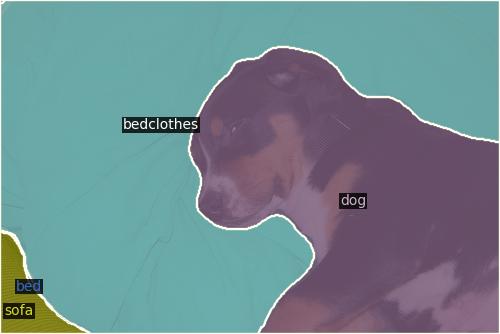} \end{subfigure}
    \begin{subfigure}[t]{0.1206\textwidth} \includegraphics[height=1.9cm,width=\textwidth]{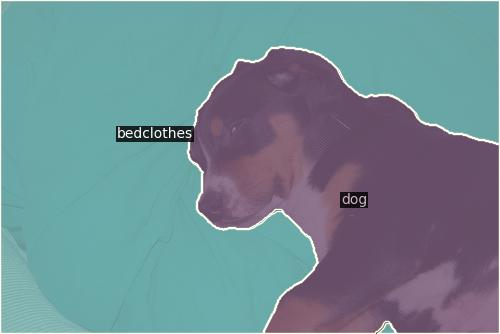} \end{subfigure}
    \begin{subfigure}[t]{0.1206\textwidth} \includegraphics[height=1.9cm,width=\textwidth]{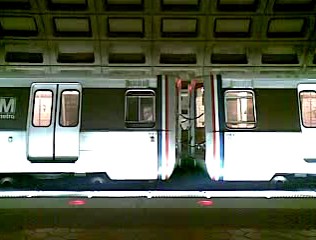} \end{subfigure}
    \begin{subfigure}[t]{0.1206\textwidth} \includegraphics[height=1.9cm,width=\textwidth]{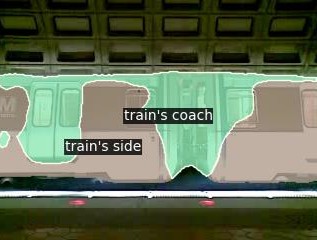} \end{subfigure}
    \begin{subfigure}[t]{0.1206\textwidth} \includegraphics[height=1.9cm,width=\textwidth]{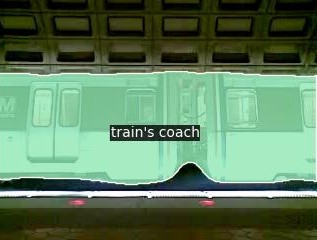} \end{subfigure}
    \begin{subfigure}[t]{0.1206\textwidth} \includegraphics[height=1.9cm,width=\textwidth]{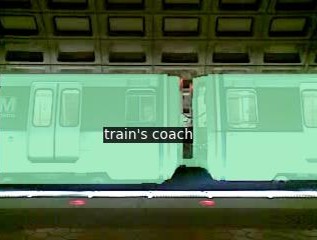} \end{subfigure}
    
    \vspace{-1em}

    \begin{subfigure}[t]{0.1206\textwidth}
        \caption{Image}
    \end{subfigure}
    \begin{subfigure}[t]{0.1206\textwidth}
   \caption{DeCLIP}
    \end{subfigure}
    \begin{subfigure}[t]{0.1206\textwidth}
\caption{Ours}
    \end{subfigure}
    \begin{subfigure}[t]{0.1206\textwidth}
        \caption{Ground-Truth}
    \end{subfigure}
    \begin{subfigure}[t]{0.1206\textwidth}
\caption{Image}
    \end{subfigure}
    \begin{subfigure}[t]{0.1206\textwidth}
    \caption{PartCATSeg }
    \end{subfigure}
    \begin{subfigure}[t]{0.1206\textwidth}
        \caption{Ours}
    \end{subfigure}
    \begin{subfigure}[t]{0.1206\textwidth}
        \caption{Ground-Truth}
    \end{subfigure}
    \vspace{-0.8em}
    \caption{
        Comparison of qualitative results with serial cost aggregation-based state-of-the-art methods  at different granularities. (a)–(d) illustrate open-vocabulary semantic segmentation on the \texttt{PC-59} dataset, while (e)–(h) present open-vocabulary part segmentation on the Pascal-Part-116 benchmark. See $\S\ref{Compare_method}$ for details.
    }
    \label{fig:vis_pred_all_qualitative}
    \vspace{-1.1em}
\end{figure*}

\noindent \textbf{Qualitative Comparison.} We present a comparative visualization of the proposed PCA-Seg and state-of-the-art methods based on serial cost aggregation for open-vocabulary semantic and part segmentation, as shown in Figure~\ref{fig:vis_pred_all_qualitative}. The visualizations clearly demonstrate that our approach consistently outperforms existing methods across a variety of challenging scenes in both semantic and part segmentation. For instance, in the semantic segmentation example, DeCLIP \cite{wang2025declip} incorrectly classifies most of the \texttt{bedclothes} region in the first row as \texttt{bed}, while in the second row, the \texttt{water} region is misclassified as \texttt{sky}. For part segmentation, PartCATSeg \cite{Choi_2025_CVPR} incorrectly identifies the \texttt{dog}'s torso as the neck and misses the nose (first row), and mislabels most of the \texttt{train}'s coach area (third row). In contrast, our method avoids these errors and achieves accurate and robust segmentation. This improvement is attributed to the parallel cost aggregation architecture, which decouples class-level semantics from spatial contextual information, and to the orthogonal regularization that facilitates learning more detailed image-text alignment cues from both knowledge streams.

\begin{table}[t]
    \begin{center}
    \renewcommand\arraystretch{1.2}
    \caption{Comparative analysis of serial aggregation and the proposed parallel aggregation. See $\S\ref{abla_studies}$ for details.}
    \vspace{-0.8em}
    {\small 
    \resizebox{\linewidth}{!}{
    \begin{tabular}{c||ccccc}
    \thickhline
      \rowcolor[rgb]{0.92,0.92,0.92}  Setting & \texttt{A-847} &   \texttt{PC-459} & \texttt{A-150}  & \texttt{PC-59}  & \texttt{PAS-20$^{b}$}  \\
        \hline \hline
        W/o Agg.   & 14.1 & 20.7 & 34.4 & 59.3 & 79.1   \\
        Serial  Agg.   & 15.3 & 21.4 & 36.3 & 60.6 & 81.3   \\
         \hdashline
      \rowcolor{unitcolor3} Parallel  Agg. & \textbf{16.1}  & \textbf{22.3} & \textbf{38.1} & \textbf{62.2} & \textbf{82.7}    \\
       
        \hline
    \end{tabular}
    }
    }
    \vspace{-14pt}
    \label{tab:compare_differnet_Agg}
    \end{center}

\end{table}

\subsection{Ablation Studies}
\label{abla_studies}

\noindent \textbf{Serial \textit{vs.} Parallel Analysis.} Table \ref{tab:compare_differnet_Agg} presents the quantitative results of the two aggregation architectures across multiple benchmarks. Without aggregation, predictions are made directly from the cost volume without refinement.
The results show that the parallel aggregation architecture consistently outperforms its serial counterpart, surpassing it by 0.9\%, 1.8\%, 1.6\%, and 1.4\% mIoU on the \texttt{PC-459}, \texttt{A-150}, \texttt{PC-59}, and \texttt{PAS-20$^b$} datasets, respectively. These improvements are attributed to the parallel aggregation design, which decouples class-level semantic learning from spatial context modeling, effectively addressing the issue of knowledge interference.

We further conduct a complexity analysis for individual serial and parallel blocks, with results in Table \ref{tab:model_com}. Each parallel block, which incorporates the EPL module and the FOD constraint, introduces only a minor overhead of 0.35 M parameters, 34.4 G FLOPs, an additional 0.96 G of GPU memory, and 0.08 s per-image latency. This modest cost is compensated by significant performance gains; for instance, PCA-Seg achieves a 3.5\% improvement in hIoU on the Pred-All setting of Pascal-Part-116 and a 1.8\% increase in mIoU on \texttt{A-150}. The complexity results, along with extensive experiments on OSPS, demonstrate that while PCA-Seg introduces acceptable computational overhead, it achieves substantial performance improvements.

\begin{table}[!t]
    \begin{center}
    \renewcommand\arraystretch{1.25}
    \caption{Complexity analysis of each serial and parallel block under the same experimental setup. See $\S\ref{abla_studies}$ for details.}
    \vspace{-0.8em}
    {\small 
    \resizebox{\linewidth}{!}{
    \begin{tabular}{c||cccc}
    \thickhline
      \rowcolor[rgb]{0.92,0.92,0.92}  Setting &  \texttt{Param(M)} & \texttt{FLOPs(G)}  & \texttt{Memory(G)}  & \texttt{Time(s)}  \\
        \hline \hline
     Entire Network & 70.3  & 2121.32  & 26.4  &  0.89   \\ 
       Serial Block & 0.68  & 53.67 & 0.60 & 0.11   \\
       \hdashline
      \rowcolor{unitcolor3} Parallel  Block & 1.03  & 88.07 & 1.56 & 0.19    \\
       
        \hline
    \end{tabular}
    }
    }
    \vspace{-20pt}
    \label{tab:model_com}
    \end{center}

\end{table}

\begin{table*}[t]
    \centering
    \caption{A set of ablative studies  on three datasets. (a) and (b) provide analyses of integrating the proposed components into the parallel and existing serial architectures, respectively. (c) and (d) show the impact of the number of expert blocks in the multi-expert parser and the weight $\lambda$ of the FOD loss on model performance, while (e) presents the performance analysis of the integration strategy for the multiple  knowledge parsed by the multi-expert (ME) parser. See $\S\ref{abla_studies}$ for details.}
    \vspace{-0.7em}
    \hspace{-0.9em}
    \renewcommand\arraystretch{1.2}
    \begin{subtable}{0.3\linewidth}
		\resizebox{\textwidth}{!}{
			\setlength\tabcolsep{2pt}
			\renewcommand\arraystretch{1.5}
        \begin{tabular}{>{\centering\arraybackslash}p{1.5cm}
        >{\centering\arraybackslash}p{1cm}>{\centering\arraybackslash}p{1cm}||>{\centering\arraybackslash}p{1.7cm}|>{\centering\arraybackslash}p{1.7cm}}
        \thickhline
        \rowcolor[rgb]{0.92,0.92,0.92}   {\fontsize{11}{20}\selectfont Baseline}  &   {\fontsize{12}{20}\selectfont EPL} & {\fontsize{11}{20}\selectfont FOD} &  {\fontsize{11}{20}\selectfont \texttt{A-150}} & {\fontsize{11}{20}\selectfont\texttt{PAS-20$^{b}$}}   \\ \hline \hline
           $\checkmark$ &  & & {\fontsize{11}{20}\selectfont 36.4} & {\fontsize{11}{20}\selectfont 81.1}  \\ \hline
           \rowcolor{table_content} $\checkmark$ & $\checkmark$  &  & {\fontsize{11}{20}\selectfont 37.2} & {\fontsize{11}{20}\selectfont 82.0} \\ \hline
            $\checkmark$ &  & $\checkmark$ & {\fontsize{11}{20}\selectfont 36.9} & {\fontsize{11}{20}\selectfont 82.1}    \\ \hline
           \rowcolor{unitcolor3} $\checkmark$ & $\checkmark$ & $\checkmark$ & {\fontsize{11}{20}\selectfont \textbf{38.1}} & {\fontsize{11}{20}\selectfont \textbf{82.7}}  \\ \hline
        \end{tabular}
		}
		\setlength{\abovecaptionskip}{0.3cm}
		\setlength{\belowcaptionskip}{-0.1cm}
		\caption{Component Analysis}
		\label{tab:component}
	\end{subtable}
    \hspace{0.001em}
    \begin{subtable}{0.16\linewidth}
		\resizebox{\textwidth}{!}{
			\setlength\tabcolsep{2pt}
			\renewcommand\arraystretch{1.32}
        \begin{tabular}{>{\centering\arraybackslash}p{1.6cm}||
        >{\centering\arraybackslash}p{1.5cm}}
        \thickhline
        \rowcolor[rgb]{0.92,0.92,0.92}  Setting  & \texttt{A-150}    \\ \hline \hline
           Serial  & 36.3  \\ \hline
           ~~~~+ EPL  &  36.6  \\ \hline
          ~~~~~+ FOD  &   36.6  \\ \hline
           \rowcolor{unitcolor3} ~~+ All  & {\normalsize\textbf{36.7}}   \\ \hline
        \end{tabular}
		}
		\setlength{\abovecaptionskip}{0.3cm}
		\setlength{\belowcaptionskip}{-0.1cm}
		\caption{Comparison Analysis}
		\label{tab:serial_analysis}
	\end{subtable}
    \hspace{0.01em}
    \begin{subtable}{0.12\linewidth}
		\resizebox{\textwidth}{!}{
			\setlength\tabcolsep{2pt}
			\renewcommand\arraystretch{1.2}
        \begin{tabular}{>{\centering\arraybackslash}p{0.9cm}||
        >{\centering\arraybackslash}p{1.5cm}}
        \thickhline
        \rowcolor[rgb]{0.92,0.92,0.92}  {\fontsize{10}{20}\selectfont $\Zcal$}  & {\fontsize{10}{20}\selectfont \texttt{A-150}}    \\ \hline \hline
           {\fontsize{10.5}{20}\selectfont 1}  & {\fontsize{10.5}{20}\selectfont 36.8}  \\ \hline
           \rowcolor{table_content}2  &  {\fontsize{10.5}{20}\selectfont 36.9}  \\ \hline
            {\fontsize{10.5}{20}\selectfont 3}  & {\fontsize{10.5}{20}\selectfont 37.0}    \\ \hline
           \rowcolor{unitcolor3} {\fontsize{10.5}{20}\selectfont \textcolor{red}{4}}  & {\fontsize{10.5}{20}\selectfont \textbf{37.2}}   \\ \hline
          {\fontsize{10.5}{20}\selectfont 5}  & {\fontsize{10.5}{20}\selectfont 36.9}    \\ \hline
        \end{tabular}
		}
		\setlength{\abovecaptionskip}{0.3cm}
		\setlength{\belowcaptionskip}{-0.1cm}
		\caption{Expert Block}
		\label{tab:expert_block}
	\end{subtable}
    \hspace{0.01em}
    \begin{subtable}{0.18\linewidth}
		\resizebox{\textwidth}{!}{
			\setlength\tabcolsep{2pt}
			\renewcommand\arraystretch{1.28}
        \begin{tabular}{>{\centering\arraybackslash}p{1.1cm}||
        >{\centering\arraybackslash}p{1.5cm}|>{\centering\arraybackslash}p{1.5cm}}
        \thickhline
        \rowcolor[rgb]{0.92,0.92,0.92} $\lambda$   & \texttt{A-150} &   \texttt{PC-59}  \\ \hline \hline
         \rowcolor{table_content}  {\fontsize{11}{20}\selectfont 0.001}  & {\fontsize{11}{20}\selectfont 37.4} & {\fontsize{11}{20}\selectfont 61.7}   \\ \hline
          {\fontsize{11}{20}\selectfont 0.005}  &  {\fontsize{11}{20}\selectfont 37.9} & {\fontsize{11}{20}\selectfont 62.1}  \\ \hline
           \rowcolor{unitcolor3} {\fontsize{11}{20}\selectfont \textcolor{red}{0.010}}  &  {\fontsize{11}{20}\selectfont \textbf{38.1}} & {\fontsize{11}{20}\selectfont \textbf{62.2}} \\ \hline
           {\fontsize{11}{20}\selectfont 0.015}  & {\fontsize{11}{20}\selectfont 38.0} & {\fontsize{11}{20}\selectfont 61.9}  \\ \hline
        \rowcolor{table_content}  {\fontsize{11}{20}\selectfont 0.020}  & {\fontsize{11}{20}\selectfont 37.8}  & {\fontsize{11}{20}\selectfont 61.6}  \\ \hline
        \end{tabular}
		}
		\setlength{\abovecaptionskip}{0.3cm}
		\setlength{\belowcaptionskip}{-0.1cm}
		\caption{FOD Loss Weight $\lambda$}
		\label{tab:weight}
	\end{subtable}
    \hspace{0.001em}
	\begin{subtable}{0.21\linewidth}
		\resizebox{\textwidth}{!}{
			\setlength\tabcolsep{2pt}
			\renewcommand\arraystretch{1.13}
        \begin{tabular}{>{\centering\arraybackslash}p{2.9cm}||
        >{\centering\arraybackslash}p{1.5cm}}
        \thickhline
       \rowcolor[rgb]{0.92,0.92,0.92} Integration Strategy   &  \texttt{PC-59}      \\ \hline \hline
       -  & 60.5 \\ \hline
          \rowcolor{table_content} Convolution & 60.7 \\ \hline
         Average & 60.8 \\ \hline
        \rowcolor{table_content}  Addition & 60.9    \\ \hline
          \rowcolor{unitcolor3} \textcolor{red}{Coefficient Mapper} & \textbf{61.2}   \\ \hline
        \end{tabular}
		}
		\setlength{\abovecaptionskip}{0.3cm}
		\setlength{\belowcaptionskip}{-0.1cm}
		\caption{ME-Parser Integration}
		\label{tab:fusion}
	\end{subtable}
    \hspace{0.01em}\\
    \label{tab_ablations}
\vspace{-18pt}
\end{table*}

\noindent \textbf{Component \& Comparison  Analysis.} Table \ref{tab:component} shows the effectiveness analysis of each component in the proposed method.  The class and spatial aggregations in DeCLIP \cite{wang2025declip} are arranged in parallel and combined with feature concatenation and a single convolution to establish the baseline.  
Integrating the expert-driven perceptual learning (EPL) module alone into the baseline results in a 0.9\% mIoU improvement on \texttt{PAS-20$^{b}$}, whereas incorporating the feature orthogonalization decoupling (FOD) strategy alone yields a 1.0\% gain. The collaborative combination of EPL and FOD achieves the highest performance improvement of 1.6\% mIoU.
These results demonstrate that the EPL module and FOD strategy enable the efficient integration of the two knowledge flows, thereby enhancing the extraction of rich image-text alignment cues from the cost volume.
Furthermore, we integrate our components into a serial architecture to highlight the importance of our  parallel design, as shown in Table \ref{tab:serial_analysis}. Integrating either EPL or FOD individually results in a slight improvement in model performance. When both components are integrated simultaneously, our proposed parallel learning paradigm outperforms the serial approach by 1.4\% mIoU, demonstrating that the proposed PCA-Seg effectively mitigates knowledge interference and captures diverse and complementary information.

\begin{figure}[t]
	\centering
	\includegraphics[width=1\linewidth]{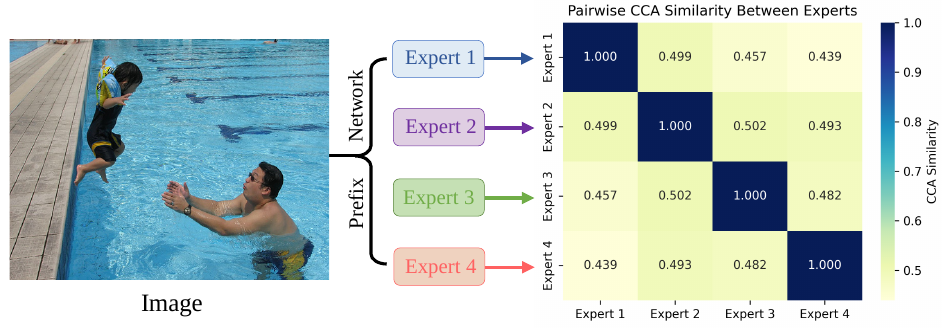}
    \vspace{-0.75cm}
	\caption{Redundancy analysis of feature knowledge among different expert blocks. See $\S\ref{abla_studies}$ for details.}
	\label{fig_redundancy_experts}
    \vspace{-0.23cm}
\end{figure}

\noindent \textbf{Parameter Analysis.} Tables \ref{tab:expert_block} and \ref{tab:weight} summarize the effects of varying the number of expert blocks $\Zcal$ and the orthogonalization decoupling loss weight $\lambda$, on model performance. Regarding the former, as the number of expert blocks increases from 1 to 4, the model's performance improves consistently across all datasets, with the best results achieved when the $\Zcal$ is set to 4. Similarly, for the latter, as the $\lambda$ increases from 0.001 to 0.010, the model exhibits a steady performance gain, reaching optimal results at $\lambda$ = 0.010. Consequently, $\Zcal$ and $\lambda$ are set to 4 and 0.010 by default in the other experiments.

\noindent \textbf{Intergation Analysis.} Table \ref{tab:fusion} presents an analysis of various feature fusion strategies for integrating multiple expert-parsed representations. Almost all integration strategies lead to performance improvements, with the proposed coefficient mapper showing the most significant improvement, achieving a 0.7\% mIoU increase. By adaptively assigning pixel-specific weights to the fine-grained features parsed by each expert, it effectively integrates diverse knowledge while preserving complementary information.

\noindent \textbf{Redundancy Analysis.} Figure~\ref{fig_redundancy_experts} presents the CCA-quantified \cite{raghu2017svcca} (canonical correlation analysis) redundancy of feature knowledge among different expert blocks. Feature redundancy among experts is mostly below 50\%, indicating that they provide rich, complementary information for improved segmentation. 

\begin{figure}[t]
	\centering
	\includegraphics[width=1\linewidth]{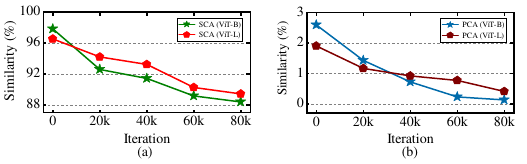}
    \vspace{-0.7cm}
	\caption{Knowledge coupling analysis between class-level semantics and spatial context under different designs. SCA/PCA denote serial/parallel cost aggregation. See $\S\ref{abla_studies}$ for details.}
	\label{fig_redundancy}
    \vspace{-0.3cm}
\end{figure}

\noindent \textbf{Knowledge Diversity Analysis.}  
We analyze the knowledge diversity between the two knowledge flows under different cost aggregation frameworks.
As shown in Figure~\ref{fig_redundancy}, the serial design exhibits significant coupling throughout training, with values remaining above 88\% even at its lowest point. This indicates that, due to knowledge interference, the two flows are inadequately disentangled in the representation space. In contrast, the proposed parallel design maintains consistently low redundancy, starting at less than 3\% and approaching zero in later stages. This demonstrates that the FOD strategy allows for more independent and informative features to be utilized by the EPL module.

\section{Conclusion}
\label{conclusion}
In this paper, we propose a \textbf{\textit{simple}} yet \textbf{\textit{effective}} parallel cost aggregation (\textbf{PCA-Seg}) paradigm designed to alleviate knowledge interference between class-level semantics and spatial context, an issue associated with the serial architecture in the OSPS domain. Specifically, we first design an expert-driven perceptual learning (EPL) module that extracts complementary feature information from both class-level semantics and spatial context, and adaptively aggregates these features using learnable pixel-specific weights. Furthermore, we further propose a feature orthogonalization decoupling (FOD) strategy to reduce redundancy between semantic and spatial information, enabling the EPL module to capture more diverse knowledge. Experimental results on eight benchmarks demonstrate that PCA-Seg achieves state-of-the-art performance.

\noindent\textbf{Acknowledgements.} This work was supported by the National Natural Science Foundation of China (No. 62506169, 62472222, U25A20442, and 62427808), Natural Science Foundation of Jiangsu Province (No. BK20240080), National Defense Science and Technology Industry Bureau Technology Infrastructure Project (JSZL2024606C001). 

{
    \small
    \bibliographystyle{ieeenat_fullname}
    \bibliography{main}
}


\end{document}